\documentclass[10pt]{article}

\usepackage{xcolor}
\usepackage{amsmath}
\usepackage{float}
\usepackage{graphicx}
\usepackage{tabularx}
\usepackage[letterpaper, top = 1.0in, bottom = 1.0in, right=0.75in, left = 0.75in]{geometry}
\usepackage{amssymb} 
\usepackage{titling}
\usepackage{caption}
\usepackage{graphicx}
\usepackage{booktabs}
\usepackage{amsmath}

\DeclareMathOperator*{\minimize}{minimize}

\usepackage{titlesec}
\usepackage{times}
\usepackage{ragged2e}
\usepackage{multirow}
\usepackage{blindtext}
\usepackage{graphics} 
\usepackage{epsfig} 
\usepackage{bm}
\usepackage{caption}
\usepackage{subcaption}
\usepackage[noadjust]{cite}
\usepackage[labelformat=simple]{subcaption}

\usepackage{dsfont}
\usepackage{relsize}
\usepackage{verbatim}
\date{}
\titleformat{\section}{\normalsize \bfseries \scshape}{\thesection}{1em}{}
\titleformat{\subsection}{\normalsize \bfseries}{\thesubsection}{1em}{}
\title{\Huge Robust GPS-Vision Localization via  \\ Integrity-Driven Landmark Attention}

\author{Sriramya Bhamidipati,~\textit{University of Illinois at Urbana-Champaign} \\
	Grace Xingxin Gao,~\textit{Stanford University} }

\begin{document}
%\RaggedRight
\maketitle
\thispagestyle{empty}

\section*{BIOGRAPHIES}
\noindent \textbf{Sriramya Bhamidipati} is a Ph.D. student in the Department of Aerospace Engineering at the University of Illinois at Urbana-Champaign, where she also received her master’s degree in 2017. She obtained her B.Tech.~in Aerospace from the Indian Institute of Technology, Bombay in 2015. Her research interests include GPS, power and control systems, artificial intelligence, computer vision, and unmanned aerial vehicles. 
~\\

\noindent \textbf{Grace Xingxin Gao} is an assistant professor in the Department of Aeronautics and Astronautics at Stanford University. Before joining Stanford University, she was an assistant professor at University of Illinois at Urbana-Champaign. She obtained her Ph.D. degree at Stanford University. Her research is on robust and secure positioning, navigation and timing with applications to manned and unmanned aerial vehicles, robotics, and power systems.

\section*{ABSTRACT}
For robust GPS-vision navigation in urban areas, we propose an Integrity-driven Landmark Attention (ILA) technique via stochastic reachability. Inspired by cognitive attention in humans, we perform convex optimization to select a subset of landmarks from GPS and vision measurements that maximizes integrity-driven performance. Given known measurement error bounds in non-faulty conditions, our ILA follows a unified approach to address both GPS and vision faults and is compatible any off-the-shelf estimator. We analyze measurement deviation to estimate the stochastic reachable set of expected position for each landmark, which is parameterized via probabilistic zonotope (p-Zonotope). We apply set union to formulate a p-Zonotopic cost that represents the size of position bounds based on landmark inclusion/exclusion. We jointly minimize the p-Zonotopic cost and maximize the number of landmarks via convex relaxation. For an urban dataset, we demonstrate improved localization accuracy and robust predicted availability for a pre-defined alert limit.

\section{INTRODUCTION} \label{sec_ION:introduction}
Sensor fusion, a widely-implemented framework for urban navigation, combines the measurements from complementary sensors, such as GPS and vision, to address the limitations of individual contributors~\cite{santoso2016visual}. These limitations induce time-varying bias in GPS and vision data that are termed as \textit{measurement faults}. For instance, GPS signals may be blocked or reflected by surrounding infrastructure, such as tall buildings and thick foliage, which significantly degrades the satellite visibility and induces multipath in the received measurements~\cite{zhu2018gnss}. In contrast, while visual odometry performs well in urban areas due to feature-rich surroundings, it suffers from data association errors due to similarities in building infrastructure, dynamic occlusions, and illumination variations~\cite{gil2006improving}. Furthermore, urban areas are susceptible to multiple measurement faults. Therefore, there is a need for integrity-driven sensor fusion that accounts for multiple faults in both GPS and vision. 

Recently, integrity is emerging as an important safety metric to assess urban navigation performance. Integrity is defined as the measure of trust that can be placed in the correctness of the computed position estimate~\cite{pullen2011sbas}. In particular, the measures of integrity that are of interest to us include position error bounds, availability, and Alert Limit~(AL), where AL is the maximum tolerable position error beyond which the system is declared to be unavailable. Traditionally, integrity is developed for GPS receivers operating in the aviation sector. These techniques utilize the redundancy in number of GPS signals available under open-sky conditions to mitigate broadcast anomalies in the satellite navigation message~\cite{hewitson2006gnss}. However, given the presence of multiple measurement faults in urban areas, applying integrity to sensor fusion-based urban navigation is not straightforward~\cite{joerger2017towards}. 

While research on GPS-vision navigation~\cite{heredia2009multi,shepard2014high,lim2017integration} is gaining momentum, to the best of our knowledge, there is only limited literature on the integrity of sensor fusion. In our prior works~\cite{bhamidipati2018GPSOnly,bhamidipati2020integrity,bhamidipati2018multisensor}, we developed a GPS-vision Simultaneous Localization And Mapping (SLAM)-based Integrity Monitoring~(IM) that performs multiple Fault Detection and Exclusion (FDE) via analysis of temporal correlation across GPS measurement residuals, and the spatial correlation across vision intensity residuals. The authors of~\cite{shytermeja2014proposed} proposed an IM framework wherein they utilized a fish-eye camera for isolating GPS faults with an assumption that the faults in vision measurements are negligible. In urban areas, there exist major challenges in addressing multiple faults in both GPS and vision, which are summarized as follows: 
\begin{itemize}
	\item 	Given the availability of a large number of measurements and the potential for multiple faults, evaluating all multi-fault hypotheses to perform FDE leads to a combinatorial explosion, and solving this NP-hard problem~\cite{woeginger2003exact} is impractical.
	\item The measurement fault distributions are unknown, multi-modal and heavy-tailed in nature. Therefore, approximating the measurement faults by known distributions~\cite{djebouri2004averaging,larson2019gaussian,rabaoui2011dirichlet}, such as Gaussian-Pareto, Rayleigh, Mixture Model etc. is not reliable enough for safety-critical applications. The need for tailored FDE approaches for each sensor, in this case GPS and vision, further limits their practical applicability.
	\item 	While FDE is important, exclusion may lead to poor Dilution Of Precision (DOP) and thereby, degrade localization accuracy~\cite{won2014selective}.
\end{itemize}

To address the above challenges, we formulate the following objectives: (a)~select a desirable subset of measurements from available GPS and vision that minimizes the position error bounds; (b)~design a unified approach to address multiple faults in GPS and vision while ensuring computational tractability; and (c)~predict integrity-driven measures of expected navigation performance for the measurement subset. This work is based on our recent ION GNSS+ $2020$ conference paper~\cite{bhamidipati2020attention}.

Our current work is inspired by the cognitive processes in humans~\cite{torralba2006contextual}. A human brain can seamlessly process large amounts of sensory cues, such as visual, auditory, olfactory, haptic and environmental, to ensure robust navigation~\cite{caduff2008assessment}. For instance, to navigate from point~A to point~B, our brain focuses on distinctive features, such as street names, stop signs, lane markings, building facade(s), etc. Furthermore, given the limited computational resources, our brain ignores other cues like clouds, dry leaves, malfunctioning road signs, and accessories of other people. Therefore, humans navigate by giving \textit{attention} to certain aspects in the surroundings, while \textit{filtering out} the irrelevant or faulty ones. 
\begin{figure}[H]
	\centering
	\begin{subfigure}[b]{0.45\textwidth}
		\includegraphics[width=\textwidth]{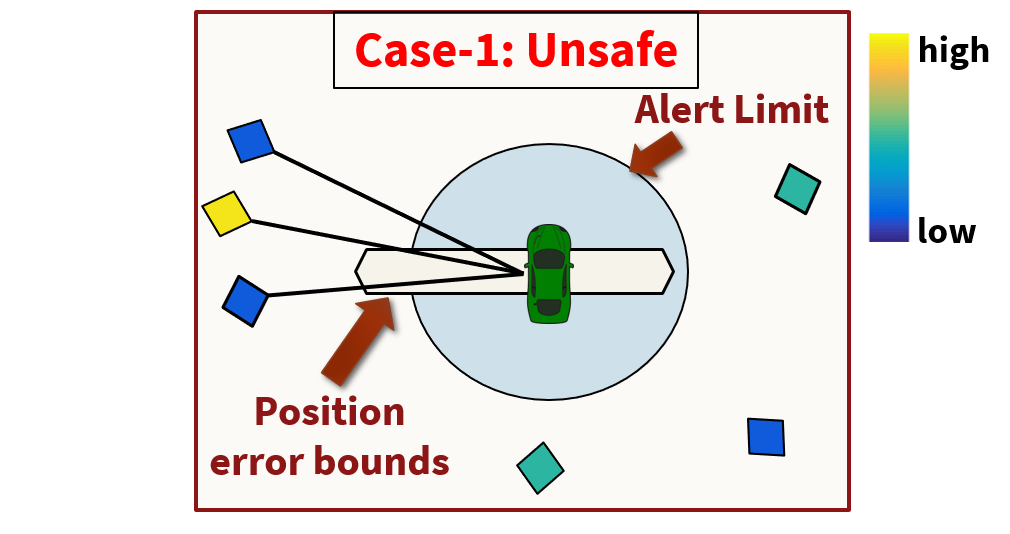}
		\caption{Unsafe}
		\label{fig_ION:one}
	\end{subfigure}	
	%\hspace{10mm}
	\begin{subfigure}[b]{0.45\textwidth}
		\includegraphics[width=\textwidth]{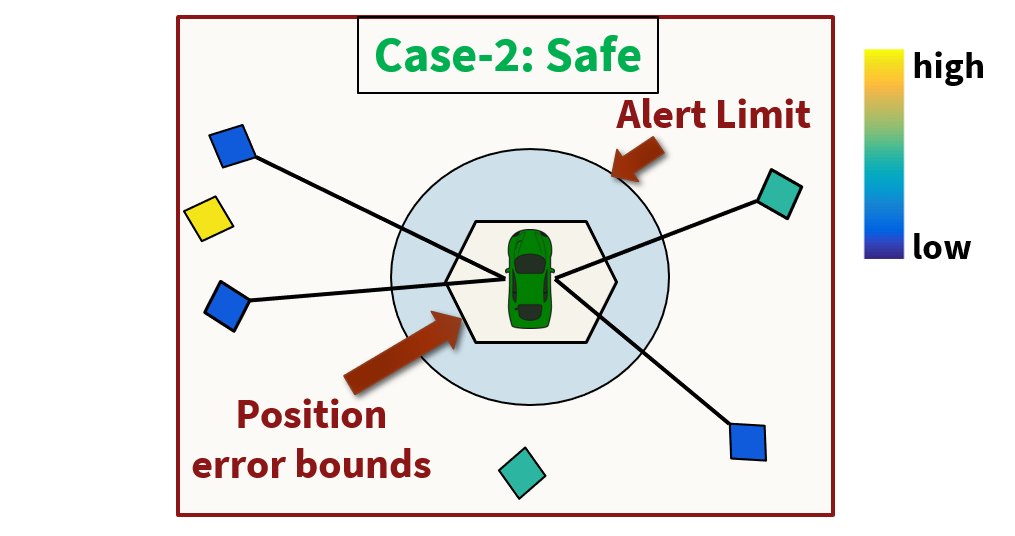}
		\caption{Safe}
		\label{fig_ION:two}
	\end{subfigure}
	\caption{Navigation performance comparison by varying the selected landmarks as follows: (a)~unsafe, as AL is violated; (b)~safe, as AL is satisfied. The blue translucent circle denotes the AL, which is pre-defined for a particular navigation task. The diamonds denote landmarks, and their measurement quality is indicated by the Parula colormap~\cite{nunez2018optimizing} that ranges from blue to yellow, with blue representing low fault magnitude and yellow to be high faults. }
	\label{fig_ION:AL}
\end{figure}
We propose an Integrity-driven Landmark Attention~(ILA) technique for GPS-vision navigation in urban areas. In this context, we define two terms: \textit{landmark} and \textit{attention}. Landmarks are the three-dimensional~(3D) features in urban surroundings that are used for navigation~\cite{lerner2007landmark}. We consider two types of landmarks: GPS satellites, whose locations are known and calculated from broadcast ephemeris~\cite{misra2006global}, and 3D visual features in urban infrastructure, whose locations are unknown and need to be triangulated via techniques such as SLAM~\cite{cadena2016past}. Attention is the process of parsing a large number of measurements from GPS and vision to select landmarks that maximize the integrity-driven navigation performance under limited computational capability. Figure~\ref{fig_ION:AL} illustrates the intuitive understanding of \textit{maximizing integrity-driven navigation performance} given the available landmarks. The subset of landmarks used to estimate the position in Fig.~\ref{fig_ION:one} includes a high measurement fault magnitude and poor DOP. This causes the estimated position error bounds to violate the pre-defined AL. In contrast, Fig.~\ref{fig_ION:two} shows a subset of desirable landmarks that exclude the necessary faults to ensure compliance of position error bounds with the AL. In this work, we maximize the integrity-driven performance of GPS-vision navigation by estimating a binary decision for each landmark that indicates whether it is included or excluded during localization. Mathematically, we represent inclusion by a binary value of $1$ and exclusion via $0$. 

\begin{figure}[H]
	\setlength{\belowcaptionskip}{-4pt}
	\centering	\includegraphics[width=0.7\columnwidth]{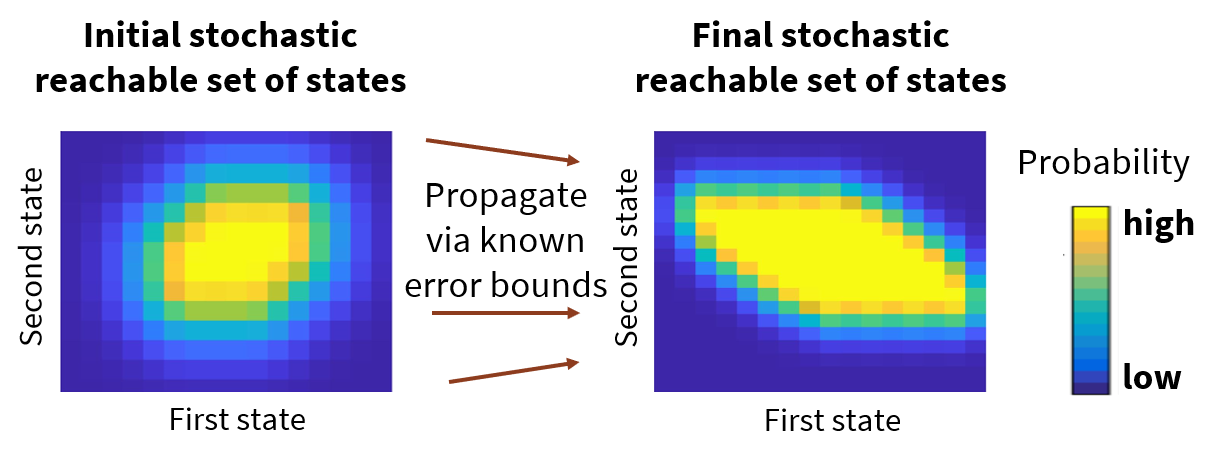}
	\caption{Illustration of SR that computes a final set of stochastic reachable set of 2D position given an initial stochastic set and known error bounds, where blue represents low probability of state vector and yellow indicates high probability.}
	\label{fig:sr_framework}
\end{figure}

In the proposed ILA technique, we utilize Stochastic Reachability~(SR) to compute not only the subset of desired landmarks, but also the predicted integrity measures, namely position error bounds and system availability, for a pre-defined AL. SR is a set-valued approach designed for stochastic processes~\cite{althoff2009safety}. It is widely used in robotics for applications such as path planning and collision avoidance, but, it has not been applied to the field of fault resilience. As seen in Fig.~\ref{fig:sr_framework}, SR computes a set of stochastic reachable states given an initial stochastic set of states, and the known stochastic sets of system disturbances and measurement errors~\cite{prandini2006stochastic}. In this context, each state in the stochastic reachable set is associated with a probabilistic measure that indicates the likelihood of its occurrence. SR accounts for the stochastic nature of error bounds associated with the motion model, GPS, and vision data, thereby making it a suitable framework for satisfying the objectives of this work.  

\subsection{Contributions}
Our ILA technique works with any off-the-shelf GPS-vision estimator, and is therefore, modular and add-on in nature. Furthermore, it follows a unified approach to account for multiple measurement faults in both GPS and vision. The main contributions of this paper are listed as follows:  
\begin{enumerate}
	\item \textbf{Landmark attention via optimization: }We design an optimization framework to select the subset of desired landmarks among those available from GPS and vision. We maximize the integrity-driven performance of GPS-vision navigation in urban areas by formulating a total cost that jointly maximizes the number of selected landmarks while minimizing the corresponding position error bounds. 
	\item \textbf{p-Zonotopic cost via SR: }Utilizing the measurement error bounds in non-faulty conditions and history of measurement residuals, we formulate the size of position error bounds using SR~\cite{prandini2006stochastic}. Specifically, we parameterize the stochastic reachable set via a representation known as the probabilistic zonotope~(p-Zonotope)~\cite{althoff2009safety}. We design the cost by formulating the p-Zonotope of position error as a union function across landmarks, wherein each landmark is associated with a binary variable that indicates its inclusion or exclusion. 
	\item \textbf{Optimization via convex relaxation: }We formulate a convex optimization problem~\cite{boyd2004convex} by relaxing the constraints on each landmark such that they lie in a continuous domain between~$[0,1]$ instead of binary domain~$\{0,1\}$. In addition, we evaluate the p-Zonotopic cost for the subset of desired landmarks to predict position error bounds and system availability. 
	\item \textbf{Experimental validation: }We have validated the performance of the proposed ILA technique on an urban sequence from a monocular visual odometry dataset~\cite{engel2016photometrically} that is combined with simulated GPS data. In the presence of multiple GPS and vision faults, the proposed ILA technique has demonstrated improved localization accuracy compared to existing methods on landmark selection and FDE. We have also computed a robust measure of the predicted system availability. 
\end{enumerate}

The rest of the paper is organized as follows: Section~\ref{sec_ION:pZonotopes} describes the preliminaries of SR and p-Zonotopes; Section~\ref{sec_ION:algorithm} describes the proposed ILA technique; Section~\ref{sec_ION:exps} experimentally demonstrates the improved localization accuracy and robust measure of predicted system availability; and Section~\ref{sec_ION:conclusions} concludes the paper.

\section{PRELIMINARIES OF SR AND p-ZONOTOPES} \label{sec_ION:pZonotopes}
\begin{figure}[H]
	\centering
	\begin{subfigure}[b]{0.3\textwidth}
		\includegraphics[width=\textwidth]{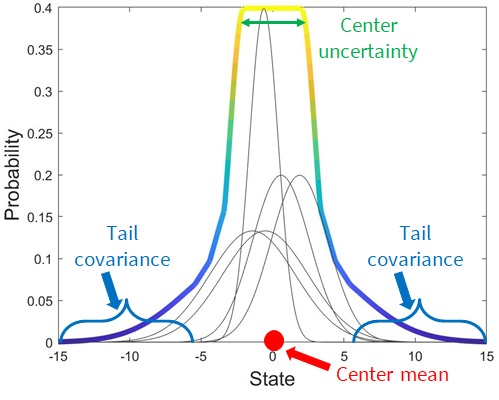}
		\caption{Example: 1D p-Zonotope}
		\label{fig_ION:oneDimEx}
	\end{subfigure}
	\hspace{10mm} %\hfill or \hspace{0.3\textwidth}
	\begin{subfigure}[b]{0.4\textwidth}
		\includegraphics[width=\textwidth]{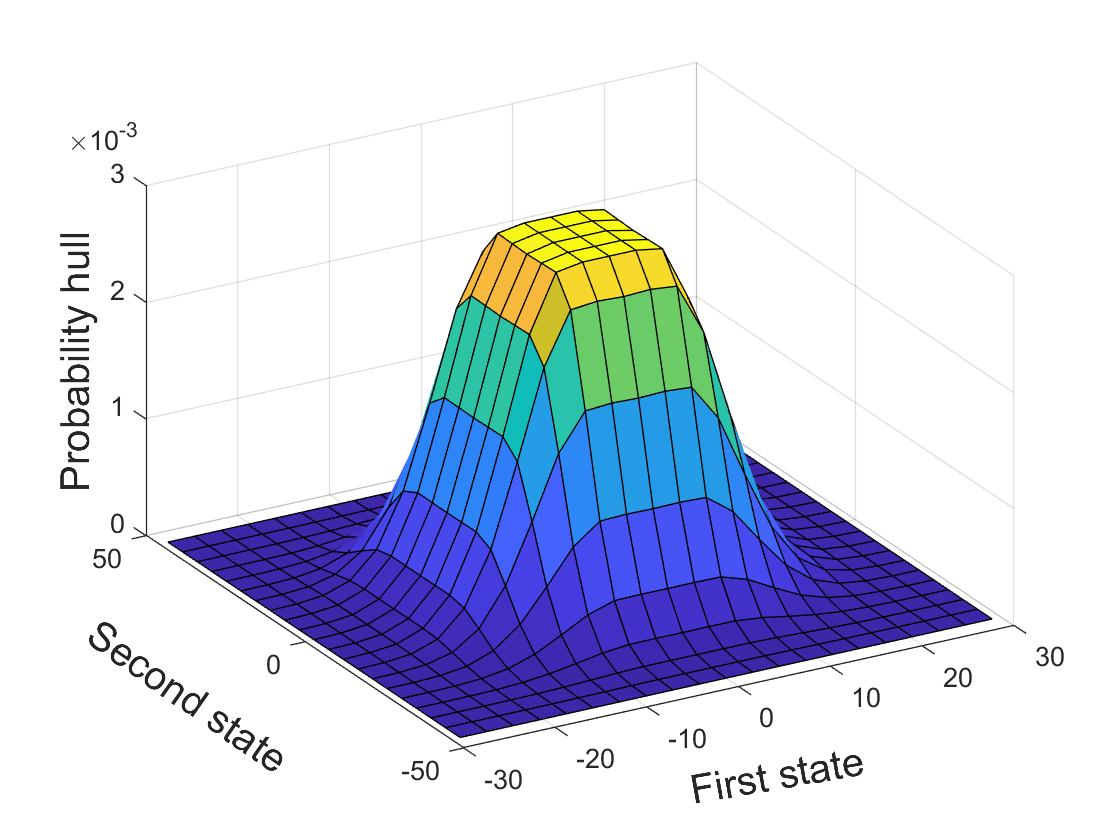}
		\caption{Example: 2D p-Zonotope}
		\label{fig_ION:twodimEx}
	\end{subfigure}
	\caption{Illustration of p-Zonotopes; (a) shows an example of a 1D p-Zonotope whose probability is indicated by the Parula colormap~\cite{nunez2018optimizing}. This 1D p-Zonotope encompasses multiple Gaussian distributions, a subset of which are shown in gray; (b) shows an example of a 2D p-Zonotope and is indicated by the Parula colormap~\cite{nunez2018optimizing}.}
\end{figure}

In this work, we represent the stochastic reachable set via a p-Zonotope. A p-Zonotope, denoted by $\mathcal{L}$, is an enclosing probabilistic hull of all possible probability density functions~\cite{althoff2009safety}. Among the existing set representations~\cite{althoff2014online} such as zonotopes, ellipsoids, polytopes, etc., we choose p-Zonotopes because they can efficiently enclose the time-varying error characteristics in GPS and vision measurements. A p-Zonotope can enclose a variety of distributions such as mixture models, distributions with uncertain or time-dependent mean, non-Gaussian or unknown distribution, etc. A one-dimensional~(1D) example of a p-Zonotope that encloses multiple Gaussian distributions with varying mean and covariance is illustrated in Fig.~\ref{fig_ION:oneDimEx} and a two-dimensional~(2D) p-Zonotope is shown in Fig.~\ref{fig_ION:twodimEx}.

A p-Zonotope $\mathcal{L}$, as seen in Eq.~\eqref{eq_ION:pZono} and Fig.~\ref{fig_ION:oneDimEx}, is characterized by
three parameters~\cite{althoff2009safety}, namely, (a)~the mean of its center, denoted by $\bm{c}\in \mathbb{R}^{n}$, (b)~the uncertainty in the center, represented by the generator matrix~$G\in \mathbb{R}^{n\times e}$, and (c)~the over-bounding covariance of the Gaussian tails, denoted by $\Sigma\in \mathbb{R}^{n\times n}$. 
\begin{subequations}
\begin{align} \label{eq_ION:pZono}
\mathcal{L}=(\bm{c},G,\Sigma)
\end{align}
Intuitively, a p-Zonotope with a certain center mean, i.e., zero center uncertainty~$G=\bm{0}$, represents a Gaussian distribution that overbounds multiple distributions with the same center mean~$\bm{c}$. This is represented by $\mathcal{Z}$ in Eq.~\eqref{eq_ION:Gzono}. 
\begin{align} \label{eq_ION:Gzono}
\mathcal{Z} = \Bigg\{\bm{c}+\sum_{i=1}^{l}\mathcal{N}^{i}(0,1)\,\underline{\bm{g}}^{i}\Bigg\},
\end{align}
\noindent where $\mathcal{N}^{i}(0,1)~\forall i\in \{1,\cdots,l\}$ denotes independent normally distributed random variables and $\underline{\bm{g}}^{i}\in \mathbb{R}^{n}$ denotes the associated generator vectors of~$\mathcal{Z}$, such that $\underline{G}=[\underline{\bm{g}}^{1},\cdots,\underline{\bm{g}}^{l}]$ and $\Sigma=\underline{G}\underline{G}^{\top}$. 

Similarly, a p-Zonotope with zero covariance~$\Sigma=\bm{0}$ simplifies to a zonotope~\cite{althoff2014online} that encompasses all the possible values of the center, such that the generator matrix of~$Z$ is $G=[\bm{g}^{1},\cdots,\bm{g}^{e}]$. This case is mathematically represented by $Z$ in Eq.~\eqref{eq_ION:zono}. %More details regarding zonotopes are found in~\cite{althoff2009safety,althoff2016cora}.
\begin{align} \label{eq_ION:zono}
Z = \Big\langle\bm{c}, G \Big\rangle=\Bigg\{\bm{c}+\sum_{i=1}^{e}\beta_{i}\,\bm{g}^{i}\Bigg| -1\leq \beta_{i}\leq 1 \Bigg\}
\end{align}
\end{subequations} 

The p-Zonotope is a combination of both sets, i.e., $\mathcal{L}=\mathcal{Z}\boxplus Z$, where $\boxplus$ is a set-addition operator, such that \newline $\mathcal{L}=\mathrm{sup} \big\{ f_{\mathcal{Z}} \Big| \mathbb{E}[\mathcal{Z}]\in Z\big\}$, where $f_{\mathcal{Z}}$ is the Probability Density Function~(PDF) of the distributions represented by $\mathcal{Z}$ and $\mathbb{E}[\cdot]$~denotes the expectation operator. From Eqs.~\eqref{eq_ION:pZono}-\eqref{eq_ION:zono}, note that  $\mathcal{L}$ depends on both $G$ and $\underline{G}$. Additionally, note that unlike a PDF, the area enclosed by a p-Zonotope does not equal to one. More details regarding the characteristics of zonotopes and p-Zonotopes can be found in the prior literature~\cite{althoff2009safety,alanwar2019distributed}. 

For example, the 2D p-Zonotope shown in Fig.~\ref{fig_ION:twodimEx} is formulated by considering the bounds on the mean and covariance of the first state to be $[-5,5]$ and $[0,2]$, respectively, and of the second state to be $[-10,10]$ and $[0,3]$, respectively. We represent this 2D p-Zonotope~$\mathcal{L}=(\bm{c}, G, \Sigma)$ by $\bm{c}=\begin{bmatrix} 0 \\ 0 \end{bmatrix}$, $G=\begin{bmatrix} 
2 & 0 \\
0 & 3 
\end{bmatrix}$ and $\Sigma=3\times\begin{bmatrix} 
2 & 0 \\
0 & 3 
\end{bmatrix}=
\begin{bmatrix} 
6 & 0 \\
0 & 9
\end{bmatrix}$. We consider a multiplication factor~$3$ in the over-bounding covariance~$\Sigma$ based on the empirical rule according to which $99.7\%$ of the values lie within three standard deviations of the mean~\cite{grafarend2006linear}.

~\\
To perform set operations on the stochastic reachable sets, we require four important properties~\cite{althoff2009safety,girard2005reachability}:  
\begin{itemize}
	\item \textit{Minkowski sum of addition: } The addition of two p-Zonotopes $\mathcal{L}_{1}=(\bm{c}_{1}, G_{1}, \Sigma_{1})$ and $\mathcal{L}_{2}=(\bm{c}_{2}, G_{2}, \Sigma_{2})$ is given by
	\begin{subequations} 
		\begin{align} \label{eq_ION:pZonoProp1}
		\mathcal{L}_{1}\bigoplus\mathcal{L}_{2} &= \big\{\bm{x}_{1}+\bm{x}_{2}\,\Big| \,\bm{x}_{1}\in \mathcal{L}_{1}, \bm{x}_{2}\in \mathcal{L}_{2}\big\}
		= (\bm{c}_{1}+\bm{c}_{2},[G_{1},G_{2}],\Sigma_{1}+\Sigma_{2}),
		\end{align}
		\noindent where~$\bigoplus$ is known as Minkowski sum operator and $[\cdot,\cdot]$ denotes a horizontal concatenation operator.
		\item \textit{Linear map: } Given a matrix $A\in \mathbb{R}^{n\times n}$ and a p-Zonotope $\mathcal{L}=(\bm{c},G,\Sigma)$, 
		\begin{align} \label{eq_ION:pZonoProp2}
		A\,\mathcal{L}&= \big\{A\bm{x}~\big|~\bm{x}\in \mathcal{L}\big\}=(A\bm{c},AG,A\Sigma A^{\top})
		\end{align}
		\item \textit{Translation: } Given a vector $\mu\in \mathbb{R}^{n}$ and a p-Zonotope $\mathcal{L}=(\bm{c},G,\Sigma)$, 
		\begin{align} \label{eq_ION:pZonoProp3}
		\mu+\mathcal{L}&=(\mu+\bm{c},G,\Sigma)
		\end{align}
		\item \textit{Union: } 
		\begin{align} \label{eq_ION:pZonoProp4}
\mathcal{L}_{1}\cup\mathcal{L}_{2} &= \big\{\kappa\bm{x}_{1}+(1-\kappa)\bm{x}_{2}\,\Big| \, \bm{x}_{1}\in \mathcal{L}_{1}, \bm{x}_{2}\in \mathcal{L}_{2}~\textrm{where}~0\leq\kappa\leq 1\big\}
\end{align}
	\end{subequations}
\end{itemize}

\section{PROPOSED ILA TECHNIQUE} \label{sec_ION:algorithm} 
\begin{figure}[h]
	\setlength{\belowcaptionskip}{-4pt}
	\centering	\includegraphics[width=0.66\columnwidth]{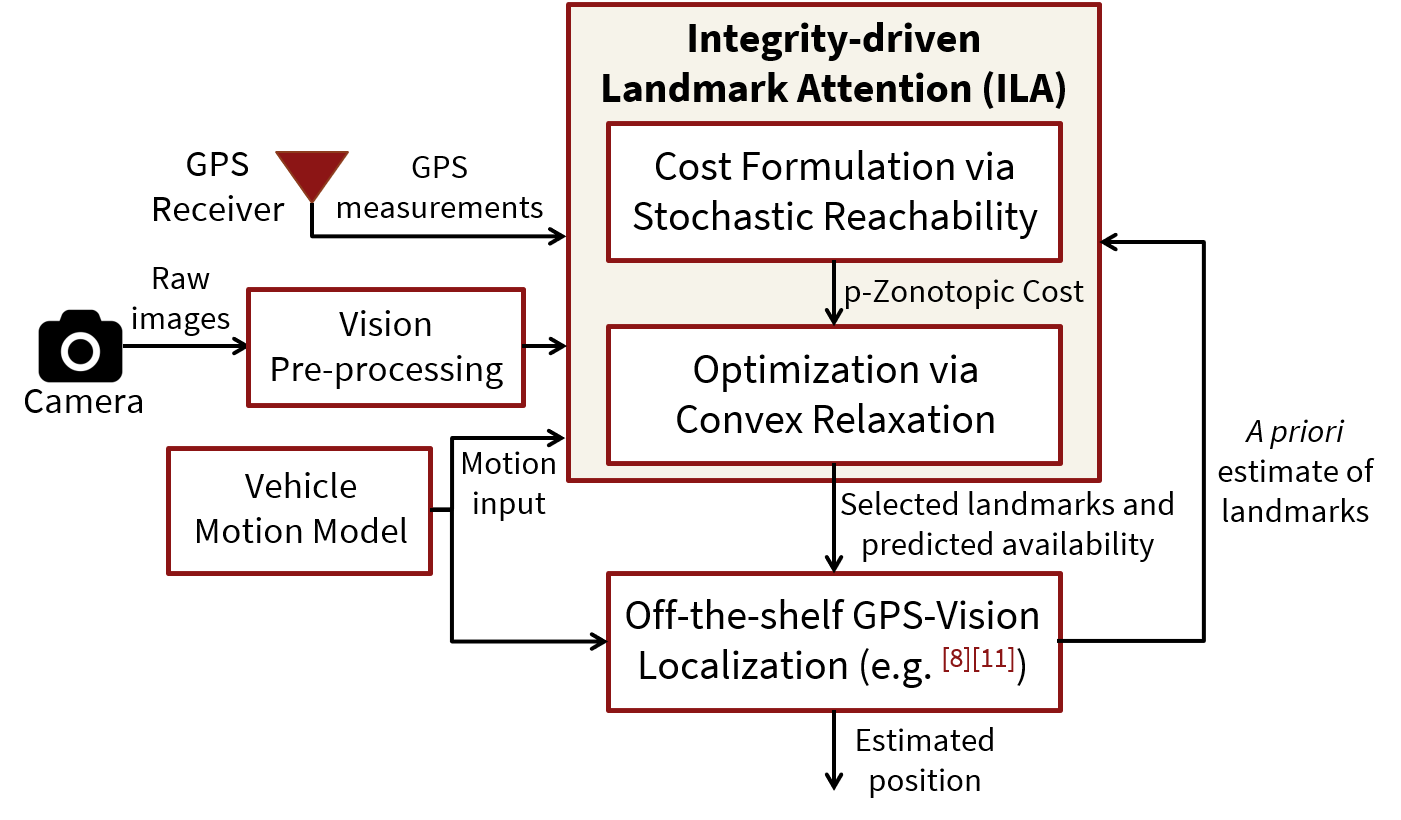}
	\caption{Architecture of the proposed ILA technique which consists of cost formulation via SR~\cite{althoff2009safety} and optimization via convex relaxation~\cite{boyd2004convex}.}
	\label{fig_ION:architecture}
\end{figure}

The proposed ILA technique shown in Fig.~\ref{fig_ION:architecture} consists of two modules: cost formulation via SR and optimization via convex relaxation. During initialization, we perform an offline empirical analysis to compute the measurement error bounds of GPS and vision in a non-faulty case. At each time iteration, the following steps are executed. 
\begin{enumerate}
	\item We consider measurements from GPS and vision, and a known motion model. We use the motion model input to facilitate global convergence of our ILA and to perform measurement fault analysis. 
	\item For each landmark, we independently analyze the deviation of received measurements against their known error bounds to estimate the p-Zonotope of expected states. Thereafter, we apply set union property of p-Zonotope to compute an over-approximated p-Zonotope that is a function of included landmarks.
	\item We jointly optimize the p-Zonotopic cost, which represents the size of over-approximated p-Zonotope, and the number of included landmarks via convex relaxation to estimate a subset of desired landmarks and a predicted measure of system availability.
	\item  The selected landmarks are later given to any off-the-shelf GPS-vision estimator module. Existing works~\cite{bhamidipati2020integrity,shepard2014high} may be referred to for details on the off-the-shelf GPS-vision estimator.
\end{enumerate}  
Denoting the number of GPS landmarks by $N$ and the number of vision landmarks by $L$, we define an attention set~$Q$ that consists of binary variables to be estimated via convex optimization, such that $Q=\Big\{q^{1}_{\mathrm{gps}},\cdots,q^{N}_{\mathrm{gps}},\,q^{1}_{\mathrm{vis}},\cdots,q^{L}_{\mathrm{vis}}\Big\}$, where $q^{i}_{\mathrm{gps}}\in\{0,1\}~\forall i\in\{1,\cdots,N\}$ and $q^{j}_{\mathrm{vis}}\in\{0,1\}~\forall j\in\{1,\cdots,L\}$. As mentioned earlier, a binary value of $1$ indicates that the landmark is included and $0$ indicates it is excluded. We formulate the total convex cost, as seen in Eq.~\eqref{eq:cost}, by jointly minimizing the p-Zonotopic cost and maximizing the number of selected landmarks. Furthermore, we define additional inequality constraints to ensure that sufficient included landmarks are available for the off-the-shelf estimator to perform GPS-vision localization. 
%\begin{subequations}
\begin{align} \label{eq:cost}
\displaystyle{\minimize_{q^{1}_{\mathrm{gps}},\cdots,q^{N}_{\mathrm{gps}},\,q^{1}_{\mathrm{vis}},\cdots,q^{L}_{\mathrm{vis}}}}\bigg(\dfrac{1}{\sum_{i=1}^{N}q_{\mathrm{gps}}^{i}+\sum_{j=1}^{L}q_{\mathrm{vis}}^{j}}\bigg)f\Big(q^{1}_{\mathrm{gps}},\cdots,q^{N}_{\mathrm{gps}},\,q^{1}_{\mathrm{vis}},\cdots,q^{L}_{\mathrm{vis}}\Big), %\bigg(\mathcal{L}\bigg),
\end{align} 
\vspace{-5mm}
\begin{align*} 
\textrm{subject~to}~ \sum_{i=1}^{N}q_{\mathrm{gps}}^{i}&>N_{\mathrm{min}},~\\
\sum_{j=1}^{L}q_{\mathrm{vis}}^{j}&>L_{\mathrm{min}},~\\
q^{i}_{\mathrm{gps}}&\in\{0,1\}~\forall i\in\{1,\cdots,N\},~\\ q^{j}_{\mathrm{vis}}&\in\{0,1\}~\forall j\in\{1,\cdots,L\},
\end{align*}
%\end{subequations} 
\noindent where $f(Q)\in \mathbb{R}$~denotes the scalar p-Zonotopic cost as a function of the attention set~$Q$ and is constructed to be convex later in Section~\ref{sec:pZonotopicCost}. This p-Zonotopic cost~$f(Q)$ depends on the p-Zonotope~$\mathcal{L}(Q)$ that represents the set union of included landmarks; $N_{\mathrm{min}}$ and $L_{\mathrm{min}}$ are the pre-defined constants that represent the minimum number of GPS and vision landmarks to be included and are the pre-defined parameters set during initialization.   

A lower p-Zonotopic cost~$f(Q)$ indicates tighter position bounds. Also, the larger the number of selected landmarks, the higher the confidence in the associated position error bounds. In the proposed ILA technique, we compute the following: the attention set~$Q^{*}$, which indicates the landmarks to be included, and the associated p-Zonotope~$\mathcal{L}(Q^{*})$, which represents the predicted position error bounds, and thereby, the predicted availability for a pre-defined AL.

\subsection{Input measurements} \label{sec:inputs}
\indent We consider the measurements from GPS, vision, and motion model as inputs to the proposed ILA technique. As explained earlier, the positions of GPS satellite landmarks are known and calculated from ephemeris~\cite{misra2006global}, while the positions of visual landmarks, i.e., 3D urban features, are unknown and localized in the off-the-shelf GPS-vision estimator. We denote the state vector at the $k$th time iteration by ${\bm x}_{k}$, such that ${\bm x}_{k}=[\textbf{x},~{\bm \psi},~c\delta t]_{k}$, where $\textbf{x}$ and ${\bm \psi}$~denote the 3D position and 3D orientation of the system with respect to a global reference frame, respectively, and $c\delta t$ denotes the clock bias of the GPS receiver. The global reference frame is set during initialization of the off-the-shelf GPS-vision estimator. Note that, the number of GPS and vision landmarks, i.e.,~$N$ and $L$, respectively, are not required to be constant across time; we drop the subscript~$k$ for notational simplicity. 

~\\
In the \textbf{GPS module}, given $N$ visible satellite landmarks, the pseudorange measurement received from the $i$th satellite at the $k$th time iteration is denoted by $z^{i}_{\mathrm{gps},k}$, such that 
\begin{align} \label{eq_ION:gps_meas_model}
z^{i}_{\mathrm{gps},k}= h^{i}\big({\bm x}_{k}\big) + f^{i}_{\mathrm{gps},k} + \eta^{i}_{\mathrm{gps},k},
\end{align}
\noindent where $h^{i}$~represents the GPS measurement model~\cite{misra2006global} associated with the $i$th satellite, with $h^{i}({\bm x}_{k})=\|\textbf{y}^{i}-\textbf{x}\|+\big(c\delta t-c\delta t^{i}\big)$; $\textbf{y}^{i}$ and $c\delta t^{i}$ denote the 3D position in global reference frame and clock corrections in the $i$th satellite, respectively; $f^{i}_{\mathrm{gps},k}$ and $\eta^{i}_{\mathrm{gps},k}$ represent the measurement fault (due to multipath) and stochastic noise associated with the $i$th satellite, respectively.

~\\
In the \textbf{vision module}, we perform direct image alignment to formulate the photometric difference between pixels across two image frames. As seen in Eq.~\eqref{eq_ION:cam_model}, we project a 2D pixel coordinate, denoted by $\textbf{u}$, from the keyframe to a current image frame at the $k$th time iteration. In this context, keyframe denotes the reference image obtained from the monocular camera relative to which the position and orientation of the current image is calculated. In a keyframe, the selected pixels with high intensity gradient are termed as key pixels. To compute the semi-dense depth map of key pixels, we execute a short-temporal baseline matching of key pixels across images to replicate the notion of stereo-matching. Detailed explanation regarding the keyframe selection and estimation of semi-dense depth maps is given in prior literature~\cite{engel2014lsd,forster2014svo}. 
\begin{align}\label{eq_ION:cam_model}
~\textrm{For}~\textbf{u}\in\Pi_{\mathrm{kf}},~I_{\mathrm{kf}}\big(\textbf{u}\big) &= I_{k}\Big(\pi\big(\omega({\bm x}_{k},\textbf{u})\big)\Big) + f_{I}(\textbf{u}) + \xi_{I}(\textbf{u}),
\end{align} 
\noindent with  
\begin{align*}
\begin{split}
\omega({\bm x}_{k},\textbf{u}) &= {\textbf R}\Big({\bm x}_{k}-{\bm x}_{\mathrm{kf}}\Big) \pi^{-1}\Big(\textbf{u},d_{\mathrm{kf}}(\textbf{u})\Big) + {\textbf t}\Big({\bm x}_{k}-{\bm x}_{\mathrm{kf}}\Big),\\
\pi(\textbf{p}) &= 
\begin{bmatrix}
\vspace{0.5pc}
f_{x} & 0 \\
0 & f_{y} 
\end{bmatrix} 
\begin{bmatrix}
\vspace{0.5pc}
\dfrac{p_{x}}{p_{z}} \\
\dfrac{p_{y}}{p_{z}}
\end{bmatrix} +
\begin{bmatrix}
c_{x}\\
c_{y}
\end{bmatrix}~\textrm{from~\cite{forster2014svo}},
\end{split}
\end{align*}
\noindent where the subscript~$\mathrm{kf}$ refers to keyframe,
\begin{enumerate}
	\item [--] $\Pi_{\mathrm{kf}} \subset \mathbb{R}^2$ and $\Pi \subset \mathbb{R}^2$ denote the domain of keyframe and current image at the $k$th time iteration, respectively. The image domain of keyframe consists of only the key pixels; %whereas current image domain comprises of all pixels; % and is set during initialization based on the resolution of the camera images;
	\item[--] $I_{\mathrm{kf}}(\textbf{u}): \Pi_{\mathrm{kf}} \rightarrow \mathbb{R}$ and $I_{k}(\textbf{u}): \Pi \rightarrow \mathbb{R}$~denote the intensity of 2D pixel coordinates~$\textbf{u}$ in the keyframe and current image at the $k$th time iteration, respectively; 
	\item [--] $\pi: \mathbb{R}^3 \rightarrow \Pi$ is the projection function that maps the 3D world coordinates, denoted by $\textbf{p}=[p_{x},p_{y}, p_{z}]^{\top}$, to a 2D pixel. From existing literature~\cite{klein2007parallel}, we define the projection function of the monocular camera via a pinhole model. We calibrate the intrinsic parameters, namely $f_{x}$, $f_{y}$, $c_{x}$, and $c_{y}$ during initialization of the off-the-shelf estimator; 
	\item [--] $\omega({\bm x}_{k},\textbf{u})$ denotes the 3D warp function that unprojects the pixel coordinates~$\textbf{u}$ and transforms it by a relative change in the state vector. This relative change is the difference between the current state vector, given by ${\bm x}_{k}$ with respect to that of keyframe, given by ${\bm x}_{\mathrm{kf}}$; 
	\item [--] ${\textbf R}(\cdot)\in \textrm{SO(3)}$ and ${\textbf t}(\cdot)\in \mathbb{R}^{3}$ denote the rotation matrix and translation vector, respectively, of the transformation between the camera frame associated with the keyframe to that of the current image frame; 
	\item [--] $\pi^{-1}(\textbf{u},d_{\mathrm{kf}}(\textbf{u})):\Pi_{\mathrm{kf}}\times \mathbb{R}^{+}\rightarrow \mathbb{R}^{3}$ denotes the unprojection function~\cite{engel2014lsd} that maps the 2D pixel~$\textbf{u}$ to 3D world coordinates via an inverse-depth, which is denoted by $d_{\mathrm{kf}}(\textbf{u})$;
	\item [--] $f_{I}(\textbf{u})$ and $\xi_{I}(\textbf{u})$ indicate the measurement fault due to data association of pixel~$\textbf{u}$ and measurement noise, respectively.
\end{enumerate}  

While we consider direct image alignment in Eqs.~\eqref{eq_ION:cam_model} and~\eqref{eq_ION:cam_meas_model}, the proposed ILA technique is also applicable to any standard feature matching framework~\cite{mur2015orb,klein2007parallel}. The number of visual landmarks is $L=|\Pi_{\mathrm{kf}}|$, where $|\Pi_{\mathrm{kf}}|$ is the cardinality of image domain~$\Pi_{\mathrm{kf}}$. From Eq.~\eqref{eq_ION:cam_model}, we formulate the non-linear vision measurement model as
\begin{align} \label{eq_ION:cam_meas_model}
z^{j}_{\mathrm{vis},k}= b^{j}\Big({\bm x}_{k},\,\textbf{p}^{j}\Big)+ f^{j}_{\mathrm{vis},k} + \xi^{j}_{\mathrm{vis},k},
\end{align}
\noindent where $\textbf{p}^{j}~\forall j\in\{1,\cdots,L\}$ represents the 3D position of the $j$th vision landmark in the global reference frame. The mapping from key pixels to the vision landmarks is bijective, i.e., $\textbf{p}^{j}=\pi^{-1}\Big(\textbf{u},d_{\mathrm{kf}}(\textbf{u})\Big)~\forall \textbf{u}\in\Pi_{\mathrm{kf}}$; $b^{j}$ represents the vision measurement model, such that $b^{j}\big({\bm x}_{k},\textbf{p}^{j}_{k}\big)=I_{k}\Big(\pi\big(\omega({\bm x}_{k},\pi(\textbf{p}^{j}))\big)\Big)$; $f^{j}_{\mathrm{vis},t}$ and $\xi^{j}_{\mathrm{vis},t}$ represent the measurement fault (due to data association) and intensity noise associated with the $j$th vision landmark in both keyframe and image frame at the $k$th iteration, respectively. 

~\\
\noindent In the \textbf{motion model}, we consider a linear state transition model to compute the state vector at the $k$th time iteration, as seen in Eq.~\eqref{eq_ION:motion_model}. Note that the proposed ILA technique is generalizable to any motion model, both linear or non-linear. Given that the focus of the proposed ILA technique is on GPS and vision faults, we consider no measurement faults in the motion model. Refer to our prior work~\cite{bhamidipati2018GPSOnly} for addressing faults in motion inputs. % instant via its motion input.
\begin{align}\label{eq_ION:motion_model}
z_{\mathrm{mm},k}&=F{\bm x}_{k-1}={\bm x}_{k} + {\bm \nu}_{k},
\end{align}
\noindent where $z_{\mathrm{mm},k}$ denotes the estimated state vector via a known motion model~$F$; ${\bm \nu}_{k}$ represents the noise vector associated with the motion model; ${\bm x}_{k-1}$ denotes the state vector computed at the previous time iteration via the off-the-shelf GPS-vision estimator, whose details are given later in Section~\ref{sec:ConvexRelaxation}. 

\subsection{p-Zonotopic cost} \label{sec:pZonotopicCost}
Utilizing the motion model, history of received measurements from GPS and vision, and their known error bounds in the non-faulty case, we formulate the p-Zonotopic cost~$f(Q_{k})$, defined earlier in Eq.~\eqref{eq:cost}, as a function of the attention set \newline $Q_{k}=\Big\{q^{1}_{\mathrm{gps},k},\cdots,q^{N}_{\mathrm{gps},k},\,q^{1}_{\mathrm{vis},k},\cdots,q^{L}_{\mathrm{vis},k}\Big\}$. We denote the known error bounds as follows: GPS measurement noise~$\eta^{i}_{\mathrm{gps},k}$ by a p-Zonotope~$\mathcal{L}^{i}_{\eta_{\mathrm{gps}},k}$, vision measurement noise~$\xi^{j}_{\mathrm{vis},k}$ by a p-Zonotope~$\mathcal{L}^{j}_{\xi_{\mathrm{vis}},k}$, and noise associated with the motion model~${\bm \nu}_{k}$ by a p-Zonotope~$\mathcal{L}_{{\bm \nu}_{\mathrm{mm}},k}$. The details regarding the estimation of measurement error bounds of GPS and vision in non-faulty conditions are given later in Section~\ref{sec_ION:exps}. We perform the following steps at the $k$th iteration: 

~\\
\noindent \textbf{Step-1: } We first formulate the set-valued position error bounds as a p-Zonotope using SR by considering non-faulty conditions for all landmarks, i.e., $f^{i}_{\mathrm{gps},k}=0~\forall i$, $f^{j}_{\mathrm{vis},k}=0~\forall j$. For this, as seen in Eqs.~\eqref{eq_ION:LinGPS_meas_model}-\eqref{eq_ION:Lin_motion_model}, we linearize the non-linear measurement models of GPS and vision that are given in Eqs.~\eqref{eq_ION:gps_meas_model}~and~\eqref{eq_ION:cam_meas_model}. Although some variants of SR explicitly handle non-linear state-space equations~\cite{august2012trajectory}, we exclude them from the scope of the current work.   
\begin{subequations}
	\begin{align} 
	\Delta z^{i}_{\mathrm{gps},k} &= H^{i}_{k}\Delta{\bm x}_{k} + \eta^{i}_{\mathrm{gps},k}~\forall i\in{1,\cdots,N}, \label{eq_ION:LinGPS_meas_model} \\[5pt]
	\Delta z^{j}_{\mathrm{vis},k} &= B^{j}_{{\bm x},k}\,\Delta{\bm x}_{k}+B^{j}_{\textbf{p},k}\,\textbf{p}^{j}_{\mathrm{apriori}} + \xi^{j}_{\mathrm{vis},k}~\forall j\in{1,\cdots,L}, \label{eq_ION:Lincam_meas_model} \\[5pt]
	\Delta z_{\mathrm{mm},k} &=\Delta{\bm x}_{k} + {\bm \nu}_{k}, \label{eq_ION:Lin_motion_model}
\end{align}	
\end{subequations} 
\noindent where $\Delta{\bm x}_{k}={\bm x}_{k}-{\bm x}_{\mathrm{apriori}}$, such that ${\bm x}_{\mathrm{apriori}}$ is the \textit{a~priori} guess of the state vector used for linearization;~$\textbf{p}^{j}_{\mathrm{apriori}}~\forall j\in\{1,\cdots,L\}$ is the \textit{a~priori} estimate of 3D visual landmarks, such that $\textbf{p}^{j}_{\mathrm{apriori}}=\bar{\textbf{p}}^{j}_{k-1}$ (assuming the visual landmarks to be stationary), where $\bar{\textbf{p}}^{j}_{k-1}$ is obtained from the off-the-shelf GPS-vision estimator; $H^{i}_{k}$~denotes the Jacobian matrix of GPS measurement model, such that $H^{i}_{k}=\dfrac{dh^{i}({\bm x}_{k})}{d{\bm x}}\Big|_{{\bm x}_{k}={\bm x}_{\mathrm{apriori}}}$; $B^{j}_{{\bm x},k}$ and $B^{j}_{\textbf{p},k}$~denote the Jacobian matrices of the vision measurement model, such that $B^{j}_{{\bm x},k}=\dfrac{db_{k}^{j}({\bm x},\textbf{p}^{j})}{d{\bm x}}\Big|_{{\bm x}={\bm x}_{\mathrm{apriori}},~\textbf{p}^{j}=\textbf{p}^{j}_{\mathrm{apriori}}}$ and $B^{j}_{\textbf{p},k}=\dfrac{db_{k}^{j}({\bm x},\textbf{p}^{j})}{d\textbf{p}^{j}}\Big|_{{\bm x}={\bm x}_{\mathrm{apriori}},~\textbf{p}^{j}=\textbf{p}^{j}_{\mathrm{apriori}}}$, respectively.

~\\
We apply set properties in Eqs.~\eqref{eq_ION:pZonoProp1}-\eqref{eq_ION:pZonoProp3} to compute the p-Zonotope of expected state estimation errors that are associated with the motion model, GPS landmarks for~$i\in\{1,\cdots,N\}$, and vision landmarks for~$j\in\{1,\cdots,L\}$. These p-Zonotopes of expected state estimation errors are defined as follows: $\mathcal{L}^{i}_{\Delta{\bm x}_{\mathrm{gps}},k}$ for the $i$th GPS landmark in Eq.~\eqref{eq_ION:OrgGPS}, $\mathcal{L}^{j}_{\Delta{\bm x}_{\mathrm{vis}},k}$ for the $j$th 3D visual landmark in Eq.~\eqref{eq_ION:OrgVision}, and $\mathcal{L}_{\Delta{\bm x}_{\mathrm{mm}},k}$ for the motion model in Eq.~\eqref{eq_ION:OrgMotion}. Note that during the non-faulty case the errors in received landmark measurements lie within the known error bounds.  

\begin{subequations}
	\begin{align} 
	\mathcal{L}^{i}_{\Delta{\bm x}_{\mathrm{gps}},k}&=\Big(\big(H^{i}_{k}\big)^{\top}H^{i}_{k}\Big)^{-1}\Big(\Delta{z}_{\mathrm{gps},k}+\mathcal{L}^{i}_{\eta_{\mathrm{gps}},k}\Big)~\forall i~\textrm{using~Eq.~\eqref{eq_ION:pZonoProp3}}, \label{eq_ION:OrgGPS} \\[10pt]
	%\end{align}
	%\begin{align} 
	\mathcal{L}^{j}_{\Delta{\bm x}_{\mathrm{vis}},k}&=\Big(\big(B^{j}_{{\bm x},k}\big)^{\top}B^{j}_{{\bm x},k}\Big)^{-1}B^{j}_{{\bm x},k}\big)^{\top}\Big( B^{j}_{\textbf{p},k}\mathcal{L}^{j}_{\textbf{p},k}\bigoplus\big(\Delta{z}_{\mathrm{vis},k}+\mathcal{L}^{j}_{\xi_{\mathrm{vis}},k}\big)\Big)~\forall j~\textrm{using~Eqs.~\eqref{eq_ION:pZonoProp1}-\eqref{eq_ION:pZonoProp3}}, \label{eq_ION:OrgVision} \\[10pt]
		%\end{align}	
	%\begin{align} 
	\mathcal{L}_{\Delta{\bm x}_{\mathrm{mm}},k}&=\Delta{\bm z}_{\mathrm{mm},k} + \mathcal{L}_{\bm{\nu}_{\mathrm{mm}},k}~\textrm{using~Eq.~\eqref{eq_ION:pZonoProp3}}, \label{eq_ION:OrgMotion} 
\end{align}	
\end{subequations} 
\noindent where $\mathcal{L}^{j}_{\textbf{p},k}$ is the p-Zonotope of the $j$th visual landmark that is formulated from its position and uncertainty estimated via the off-the-shelf GPS estimator. Observing Eqs.~\eqref{eq_ION:OrgGPS}-\eqref{eq_ION:OrgVision}, note that compared to GPS landmarks, vision landmark formulation has an additional p-Zonotope term indicating its localization uncertainty.  

\begin{figure}[h]
	\centering
	\begin{subfigure}[b]{0.4\textwidth}
		\includegraphics[width=\textwidth]{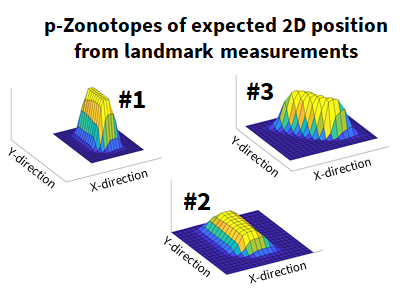}
		\caption{p-Zonotope of expected 2D positions}
		\label{fig_ION:SetUnionOne}
	\end{subfigure}
	\hspace{10mm} 
	\begin{subfigure}[b]{0.4\textwidth}
		\includegraphics[width=\textwidth]{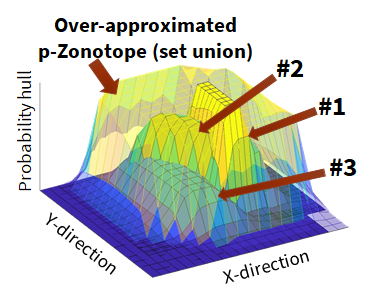}
		\caption{Over-approximated p-Zonotope}
		\label{fig_ION:SetUnionTwo}
	\end{subfigure}
	\caption{An example illustrating the set union operation and over-approximation; (a)~shows p-Zonotopes of expected 2D position from three landmarks, and (b)~shows the over-approximated 2D p-Zonotope that encloses the union of the p-Zonotopes.  }
	\label{fig_ION:pZonotopicCost}	
\end{figure}

\noindent \textbf{Step-2: } We utilize the set union property of p-Zonotopes, defined earlier in Eq.~\eqref{eq_ION:pZonoProp4}, to combine the p-Zonotopes of expected states from the motion model, as well as desired GPS and vision landmarks. Therefore, the set union of p-Zonotopes, as seen in Eq.~\eqref{eq_ION:union}, is a linear function of the attention set~$Q_{k}$. Note that the state estimation error in GPS-vision localization depends on the associated off-the-shelf estimator module described later in Section~\ref{sec_ION:exps}. The set union of p-Zonotopes computes a conservative position error bound that is valid for any off-the-shelf GPS-vision module. Given that set union is not closed under union, we over-approximate the union of p-Zonotopes via a p-Zonotope~\cite{le2010reachability}. We denote the over-approximated p-Zonotope in Eq.~\eqref{eq_ION:union} by~$\mathcal{L}(Q)$. For an intuitive understanding of set union, Fig.~\ref{fig_ION:SetUnionOne} shows an example of p-Zonotopes of expected 2D position from three landmarks~$\#1$ to $\#3$, and Fig.~\ref{fig_ION:SetUnionTwo} demonstrates the set union of these three p-Zonotopes that is over-approximated by another p-Zonotope. During implementation, we utilize the \textit{enclose} function of the MATLAB COntinuous Reachability Analyzer~(CORA) toolbox~\cite{althoff2016cora} to perform set union operation on p-Zonotopes. Note that Eq.~\eqref{eq_ION:union} considers the measurement errors to lie within the known error bounds, but, this assumption is invalid during the presence of faults. 
\begin{align} \label{eq_ION:union}
\mathcal{L}(Q_{k}) = \mathcal{L}_{\Delta{\bm x}_{\mathrm{mm}},k}\bigcup\limits_{i\in N} q_{\mathrm{gps}}^{i}\mathcal{L}^{i}_{\Delta{\bm x}_{\mathrm{gps}},k}\,\bigcup\limits_{j\in L} q_{\mathrm{vis}}^{j}\mathcal{L}^{j}_{\Delta{\bm x}_{\mathrm{vis}},k}
\end{align}

\noindent \textbf{Step-3: } We design a unified approach to account for GPS and vision measurement faults by leveraging the stochastic nature of the p-Zonotope. We utilize the received measurements and motion model to estimate the measurement innovation for GPS and vision landmarks as $\bm{\epsilon}^{i}_{\mathrm{gps}} = \Delta{\bm z}^{i}_{\mathrm{gps}}-
H^{i}\Delta{\bm x}_{\mathrm{mm}}$ and $\bm{\epsilon}^{j}_{\mathrm{vis}} = \Delta{\bm z}^{j}_{\mathrm{vis}}-B^{j}_{\bm x}\,\Delta{\bm x}_{\mathrm{mm}}-B^{j}_{\textbf{p}}\,\textbf{p}^{j}_{\mathrm{apriori}}$, respectively, with \newline $\Delta{\bm x}_{\mathrm{mm}}={\bm z}_{\mathrm{mm}}-{\bm x}_{\mathrm{apriori}}$. Utilizing the p-Zonotope of expected states from motion model~$\mathcal{L}_{\Delta{\bm x}_{\mathrm{mm}}}$ defined earlier in Eq.~\eqref{eq_ION:OrgMotion}, we apply the set properties of p-Zonotopes in Eqs.~\eqref{eq_ION:pZonoProp1}-\eqref{eq_ION:pZonoProp2} on the measurement innovation to compute the p-Zonotope of expected measurement innovation for GPS and vision landmarks as follows:
\begin{align} \label{eq_ION:pZonores}
\mathcal{L}^{i}_{\epsilon_{\mathrm{gps}}} &= H^{i}\mathcal{L}_{\Delta{\bm x}_{\mathrm{mm}}}\bigoplus\mathcal{L}^{i}_{\eta_{\mathrm{gps}}} \\
\mathcal{L}^{j}_{\epsilon_{\mathrm{vis}}} &= B^{j}_{{\bm x},k}\mathcal{L}_{\Delta{\bm x}_{\mathrm{mm}}}\bigoplus B^{j}_{\textbf{p},k}\mathcal{L}^{j}_{\textbf{p}_{\mathrm{apriori}}}\bigoplus\mathcal{L}^{j}_{\xi_{\mathrm{vis}}}
\end{align}

For each landmark, we independently analyze the deviation of estimated measurement innovation from the p-Zonotope of expected measurement innovation that is indicative of non-faulty measurement errors. The p-Zonotope of expected measurement innovation provides a probability value corresponding to each estimated measurement innovation and this probability is denoted by~$\alpha^{i}_{\mathrm{gps}}~\forall i\in\{1,\cdots,N\}$ for GPS landmarks and~$\alpha^{j}_{\mathrm{vis}}~\forall j\in\{1,\cdots,L\}$ for vision landmarks. Based on this, for each available landmark, we independently compute the fault status of received measurements by normalizing the obtained probability value with the probability value corresponding to the center mean of the p-Zonotope and subtracting it from one. Each measurement fault status~$\tilde{\alpha}^{i}_{\mathrm{gps}}\in \mathbb{R}~\forall i$ and $\tilde{\alpha}^{j}_{\mathrm{vis}}\in \mathbb{R}~\forall j$ lies between $[0,1]$. In a non-faulty condition, the estimated measurement innovation and its p-Zonotope of expected measurement innovation are in close agreement, and hence a low fault status~$\approx 0$ is obtained. However, in the presence of measurement faults, the estimated measurement innovation does not comply with this expected p-Zonotope, and therefore a high value of fault status~$\approx 1$ is observed.  

We utilize the history of past $K$ measurements to compute a joint fault status that represents the temporal confidence in the landmark. This also satisfies the need for temporal measurements of a visual landmark to triangulate its unknown position. At the $k$th time iteration, for each GPS landmark~$i\in\{1,\cdots,N\}$, we utilize~${\bm z}^{i}_{\mathrm{gps},K}=\big\{z^{i}_{\mathrm{gps},k-K},\cdots,z^{i}_{\mathrm{gps},k}\big\}\in \mathbb{R}^{K}$ to compute the corresponding joint fault status~$\alpha^{i}_{\mathrm{gps},k}\in \mathbb{R}$, and for each visual landmark~$j\in\{1,\cdots,N\}$, we utilize \newline ${\bm z}^{j}_{\mathrm{vis},K}=\big\{z^{j}_{\mathrm{vis},k-K},\cdots,z^{j}_{\mathrm{vis},k}\big\}\in \mathbb{R}^{K}$ to compute the corresponding joint fault status~$\alpha^{j}_{\mathrm{vis},k}\in \mathbb{R}$. We utilize the inverse of one minus estimated fault status for each landmark to adaptively scale the size of non-faulty measurement error bounds. Given that the size of set union increases with the size of scaled measurement error bounds, intuitively, landmarks with fault status~$\approx 0$ are more desirable to be included. Thereafter, we modify the union formulation of p-Zonotopes in Eq.~\eqref{eq_ION:union} to adaptively account for measurement faults, as seen in Eq.~\eqref{eq_ION:ModiUnion}.    
\begin{align} \label{eq_ION:ModiUnion}
\mathcal{L}(Q_{k}) = \mathcal{L}_{\Delta{\bm x}_{\mathrm{mm}},k}\bigcup\limits_{i\in N} \Big(q_{\mathrm{gps}}^{i}\big(1-\tilde{\alpha}^{i}_{\mathrm{gps},k}\big)^{-1}\mathcal{L}^{i}_{\Delta{\bm x}_{\mathrm{gps}},k}\Big)\,\bigcup\limits_{j\in L} \Big(q_{\mathrm{vis}}^{j}\big(1-\tilde{\alpha}^{j}_{\mathrm{vis},k}\big)^{-1}\mathcal{L}^{j}_{\Delta{\bm x}_{\mathrm{vis}},k}\Big)
\end{align}

\begin{figure}[h]
	\setlength{\belowcaptionskip}{-4pt}
	\centering	\includegraphics[width=0.45\columnwidth]{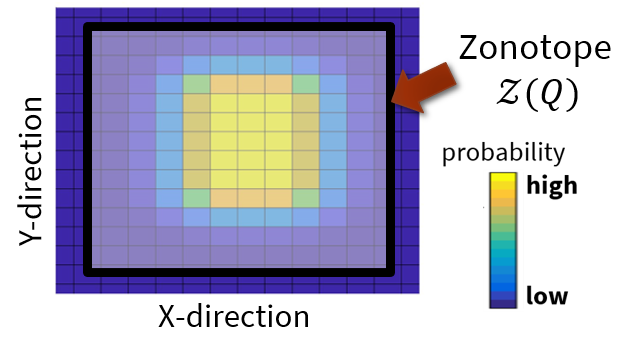}
	\caption{Illustration representing the top view of an over-approximated p-Zonotope of expected 2D position, denoted by a Parula colormap~\cite{nunez2018optimizing}. The over-approximated p-Zonotope is represented by a zonotope~$Z(Q)$ in white. }
	\label{fig_ION:Transform}
\end{figure}

\noindent \textbf{Step-4: } We formulate the scalar p-Zonotopic cost~$f(Q_{k})$ defined earlier in Eq.~\eqref{eq:cost}, to denote the size of the over-approximated p-Zonotope~$\mathcal{L}(Q_{k})$. For computing a finite size of the p-Zonotope, we cut-off the over-approximated p-Zonotope $\mathcal{L}(Q_{k})$ based on a $\gamma$-confidence set, an example of which is seen in Fig.~\ref{fig_ION:Transform}. The $\gamma$-confidence set~\cite{althoff2009safety} thresholds the tail of a p-Zonotope by a pre-defined parameter~$\gamma$, such that $P[-\gamma\leq\mathcal{N}(0,1)\leq\gamma]=\textrm{erf}(\dfrac{\gamma}{\sqrt{2}})$, where $\textrm{erf}$ denotes the error function~\cite{oldham2008error}. Note that $\gamma$ is analogous to the term \textit{integrity risk} commonly used in the GNSS community~\cite{misra2006global}. 

The cut-off p-Zonotope is represented by a zonotope, denoted by~$Z(Q_{k})$. Based on Eq.~\eqref{eq_ION:zono}, we define \newline $Z(Q_{k})=\big\langle c_{Z}(Q_{k}),G_{Z}(Q_{k}) \big\rangle$, where $c_{Z}(Q_{k})$ and $G_{Z}(Q_{k})$ denotes the center and generator matrix of the zonotope, respectively. The size of the zonotope~$Z(Q_{k})$ is given by $G_{Z}^{\top}(Q_{k})G_{Z}(Q_{k})$~\cite{girard2005reachability}. As seen in Eq.~\eqref{eq:pZonoCost}, we define p-Zonotopic cost as the weighted sum of eigenvalues of $G_{Z}^{\top}(Q_{k})G_{Z}(Q_{k})$ along each axis. We perform weighted sum to account for the perturbation sensitivity of the state estimation error across each axis. The weights also account for the linearization-based approximation of measurement models considered to apply SR in \textbf{Step-1}. 
\begin{align} \label{eq:pZonoCost}
f(Q_{k}) = \sum_{n=1}^{M}w_{n}\lambda_{n}\Big(G_{Z}^{\top}(Q_{k})G_{Z}(Q_{k})\Big),
\end{align}
\noindent where $\lambda_{n}$ denotes the eigenvalue of~$G_{Z}^{\top}(Q_{k})G_{Z}(Q_{k})$ and $w_{n}$ represents the associated weights, such that~$\sum_{n=1}^{M}w_{n}=1$ and $M$ denotes the number of states in the vector~${\bm x}_{k}$.  

\subsection{Optimization via Convex Relaxation} \label{sec:ConvexRelaxation}
We utilize the p-Zonotopic cost derived in Eq.~\eqref{eq:pZonoCost} to formulate the binary convex optimization problem represented earlier in Eq.~\eqref{eq:cost}. However, given the binary constraints on the attention set~$Q_{k}$, computing the exact solution is still an NP-hard problem. Therefore, we perform convex relaxation~\cite{carlone2018attention} in which the binary constraints of the attention set~$\{0,1\}$ are replaced by convex constraints that lie in continuous domain between~$[0,1]$. The complete optimization problem of the proposed ILA technique via convex relaxation is given by  
\begin{align} \label{eq:costModi}
\displaystyle{\minimize_{Q=\Big\{q^{1}_{\mathrm{gps}},\cdots,q^{N}_{\mathrm{gps}},\,q^{1}_{\mathrm{vis}},\cdots,q^{L}_{\mathrm{vis}}\Big\}}}\bigg(\dfrac{1}{\sum_{i\in N}q_{\mathrm{gps}}^{i}+\sum_{j\in L}q_{\mathrm{vis}}^{j}}\bigg)f\Big(q^{1}_{\mathrm{gps}},\cdots,q^{N}_{\mathrm{gps}},\,q^{1}_{\mathrm{vis}},\cdots,q^{L}_{\mathrm{vis}}\Big)
\end{align}
\vspace{-5mm}
\begin{align*} 
\textrm{subject~to} \sum_{i=1}^{N}q_{\mathrm{gps}}^{i}&>N_{\mathrm{min}},~\\
\sum_{j=1}^{L}q_{\mathrm{vis}}^{j}&>L_{\mathrm{min}},~\\
q^{i}_{\mathrm{gps}}&\in [0,1]~\forall i\in\{1,\cdots,N\},~\\ q^{j}_{\mathrm{vis}}&\in[0,1]~\forall j\in\{1,\cdots,L\}.
\end{align*}

We utilize the MATLAB CVX toolbox~\cite{grant2014cvx} to perform the above convex optimization and to estimate the attention set~$Q^{*}$. Given that the solution is not binary, we perform a simple rounding procedure, such that the landmarks above a pre-defined threshold~$\beta$ are included while the others are excluded. This pre-defined threshold~$\beta$ is set during initialization. This final rounded attention set is denoted by~$Q^{\circ}$. Note that other sophisticated methods to perform the rounding procedure are found in existing literature~\cite{lerner2007landmark,carlone2018attention}. 

From the rounded attention set~$Q^{\circ}$, the landmarks to be included are identified and used to estimate the value of $f\big(\mathcal{L}(Q^{\circ}_{k})\big)$, which represents the size of the predicted position bounds. By comparing these position bounds against a pre-defined AL, we compute the predicted availability. 

\section{EXPERIMENTAL RESULTS} \label{sec_ION:exps}
We validate the robustness of the proposed ILA technique that not only accounts for multiple GPS and vision faults in a unified manner, but also predicts the expected navigation performance.  
\begin{figure}[h]
	\setlength{\belowcaptionskip}{-4pt}
	\centering	\includegraphics[width=0.6\columnwidth]{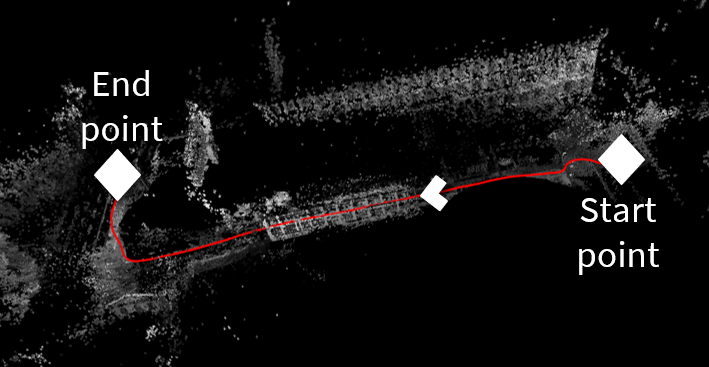}
	\caption{Trajectory visualization for an experiment time duration of $60~$s. This represents an urban sequence from the monocular visual odometry dataset~\cite{engel2016photometrically}. The surroundings include a narrow alleyway, open-sky and tall buildings, therefore, the GPS and vision measurements are susceptible to faults at multiple instances. }
	\label{fig_ION:exp_setup}
\end{figure}

\subsection{Implementation Details}
We consider an urban sequence from the monocular visual odometry dataset~\cite{engel2016photometrically} that comes with a high-fidelity reference ground truth and camera calibration.
Figure~\ref{fig_ION:exp_setup} shows the trajectory visualization for an experimental duration of $60$~s, wherein the surroundings include a narrow alleyway, open-sky, and many tall buildings. As explained in Section~\ref{sec:inputs}, we pre-process the image frames via direct image alignment~\cite{forster2014svo} to extract key pixels in each frame. To adapt this urban dataset to our input measurements, we utilize the ground truth 3D position and online-available satellite ephemeris to simulate GPS data. Particularly, we use a C++ language-based software-defined GPS simulator known as GPS-SIM-SDR~\cite{gpssdrsim,bhamidipati2019gps} to generate the raw GPS signals. Later, we post-process the simulated GPS signals using a MATLAB-based software-defined radio known as SoftGNSS~\cite{softGNSS}. When navigating through a narrow alleyway, i.e., during $t=9-24~$s, we induce simulated GPS faults, i.e., multipath and satellite blockage, based on the elevation and azimuth of the satellites with respect to the direction of travel. We add simulated multipath effects in the low-elevation GPS satellites, i.e., $20^{\circ}-45^{\circ}$, and within the azimuth range of $45^{\circ}-135^{\circ}$ and $225^{\circ}-315^{\circ}$. Similarly, for the same range of azimuth, we block the GPS satellites with elevation angles $<20^{\circ}$. Based on existing literature~\cite{phelts2000narrowband}, we induce multipath errors that range between $20-65~$m. Given that there are no standardized safety metrics as of September $2020$, we set the pre-defined $AL=7.5~$m based on the general street specifications.

We perform integrity-driven convex optimization to compute a subset of desired landmarks using the open-source MATLAB CVX toolbox~\cite{grant2014cvx}. 
We perform the set operations of SR and transform across various set representations, such as polytopes, zonotopes, p-Zonotopes, etc., using the open-source MATLAB CORA toolbox~\cite{althoff2016cora}. We initialize the proposed ILA technique by pre-defining the measurement error bounds in non-faulty conditions. For each GPS satellite landmark, we define the simulated measurement errors~$\eta^{i}_{\mathrm{gps}}$ observed during authentic conditions to have the following characteristics: mean and covariance of the time-varying Gaussian distribution lies between [$-5~$m,\,$5~$m] and [$0~$m,\,$5~$m], respectively. We formulate the p-Zonotopes from above-listed error bounds in the same manner as the example discussed earlier in Section~\ref{sec_ION:pZonotopes}. We perform empirical analysis on the estimated covariance~\cite{engel2014lsd} during multiple urban sequences from the monocular visual odometry dataset~\cite{engel2016photometrically} to compute the upper bounds on the vision measurement errors~$\xi^{i}_{\mathrm{vis}}$. Note that the measurement errors in non-faulty conditions largely depend on the type of GPS and vision sensor utilized, and therefore a generalized approach for their robust analysis is beyond the scope of this work. 
We leverage our prior work~\cite{bhamidipati2018multisensor} for designing the off-the-shelf GPS-vision estimator. We consider a history of past measurements, such that $K=8$, for both the proposed ILA technique and off-the-shelf GPS-vision estimator. If the predicted system availability at $(k-1)$th time iteration is~$1$, then we update the motion model~${\bm x}_{k-1}$ seen in Eq.~\eqref{eq_ION:motion_model} using the state vector estimated via the off-the-shelf GPS-vision estimator, otherwise~${\bm x}_{k-1}={\bm z}_{\mathrm{mm},k-1}$. We pre-define the following parameters based on heuristic analysis as follows: $\beta=0.75$, $N_{\mathrm{min}}=5$, $L_{\mathrm{min}}=600$ and $\gamma=0.999$. 

\subsection{Results and Discussion}
In Fig.~\ref{fig_ION:positionerr}, we demonstrate an improved 2D position accuracy via the proposed ILA technique as compared to other subsets of landmarks, namely only GPS landmarks, all GPS and vision landmarks, and random selection. In the random selection technique, maintaining the number of GPS and vision landmarks to be the same as that of the proposed ILA technique, we run $300$ Monte Carlo runs and compute the minimum exhibited localization error at each instant. As seen in Table~\ref{table:postn}, the proposed ILA technique demonstrates a small maximum error of only $6.6~$m compared to other techniques that violate AL as their associated maximum position errors are $>13~$m. 

\begin{table}
	{\centering
	\begin{minipage}{0.3\columnwidth}
		\caption{Quantitative comparison of maximum position error for different subsets of landmarks }
		\label{table:postn}
		\centering
		\begin{tabular}{lrr}
			\toprule
			\textbf{Landmarks Selected}         & \textbf{Max Error} \\
			\midrule
			Proposed ILA     & $6.6~$m     \\
			Only GPS  & $20.2~$m     \\
			All GPS and vision     & $13.9~$m    \\
			Random (300 runs) & $25.2~$m    \\
			\bottomrule
		\end{tabular}
	\end{minipage}\hfill
	\begin{minipage}{0.65\columnwidth}
		\centering
		\includegraphics[width=0.9\columnwidth]{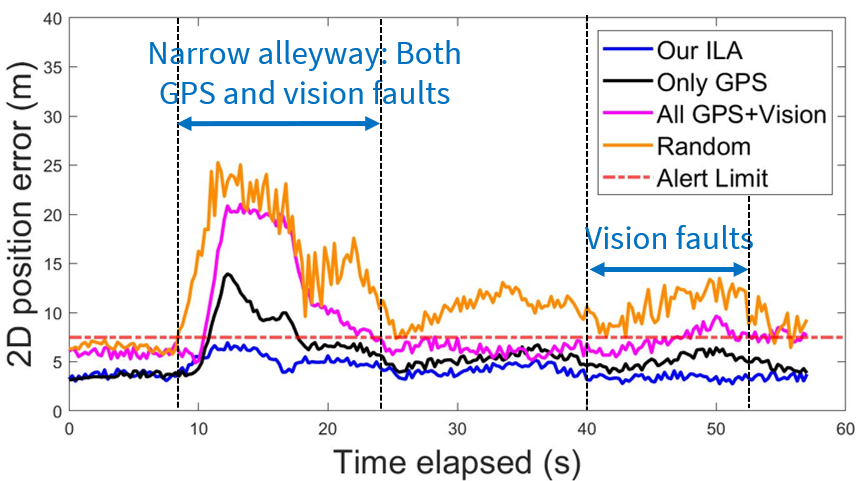}
		\captionof{figure}{Qualitatitve comparision of maximum position error when the following landmark subsets are selected: only GPS landmarks, indicated by black; all GPS and vision landmarks, represented by magenta; the proposed ILA technique, denoted by blue; and random selection, indicated in orange. Our ILA shows a smaller localization error as compared to other methods. }
		\label{fig_ION:positionerr}
	\end{minipage}
}
\end{table}

At a time instant $t=54~$s, we compare the statistics of accuracy and the number of landmarks across the above-mentioned four techniques for landmark selection. Compared to other methods seen in Table~\ref{table:landmarks}, a low accuracy of $2.9~$m is achieved using our ILA technique that selects $6$ out of $8$ GPS landmarks and $956$ out of $4722$ visual landmarks. Fig.~\ref{fig_ION:landmarks} depicts the selected landmarks in GPS and vision at time~$t=54~$s. In addition, our ILA technique successfully bounds the estimated position error wherein the predicted position error bound is $5.1~$m. Therefore, the system is predicted to be available. 
\begin{table}[H]
	{\centering
		\begin{minipage}{0.45\columnwidth}
			\caption{Comparison of maximum position error and number of landmarks at $t=54~$s across different landmark selection methods with $G:$ GPS landmarks, and $V:$ vision landmarks}
			\label{table:landmarks}
			\centering
			\begin{tabular}{lrr}
				\toprule
				\textbf{Methods}  & \textbf{Max Error} & \textbf{landmarks}\\
				\midrule
				Proposed ILA     & $2.9~$m   & $6~(G), 956~(V)$   \\
				Only GPS  & $6.9~$m   & $6~(G), 956~(V)$  \\
				All GPS and vision     & $5.4~$m   & $6~(G), 956~(V)$ \\
				Random (300 runs) & $9.1~$m  & $6~(G), 956~(V)$   \\
				\bottomrule
			\end{tabular}
		\end{minipage}\hfill
		\begin{minipage}{0.55\columnwidth}
			\centering
			\includegraphics[width=0.9\columnwidth]{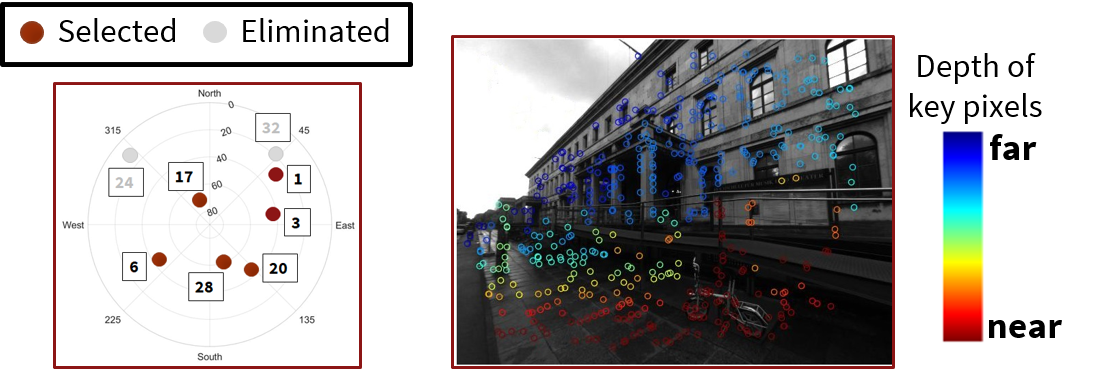}
			\captionof{figure}{Selected GPS and vision landmarks via the proposed ILA technique at $t=54~$s.}
			\label{fig_ION:landmarks}
		\end{minipage}
	}
\end{table}

\begin{figure}[h]
	\setlength{\belowcaptionskip}{-4pt}
	\centering	\includegraphics[width=0.5\columnwidth]{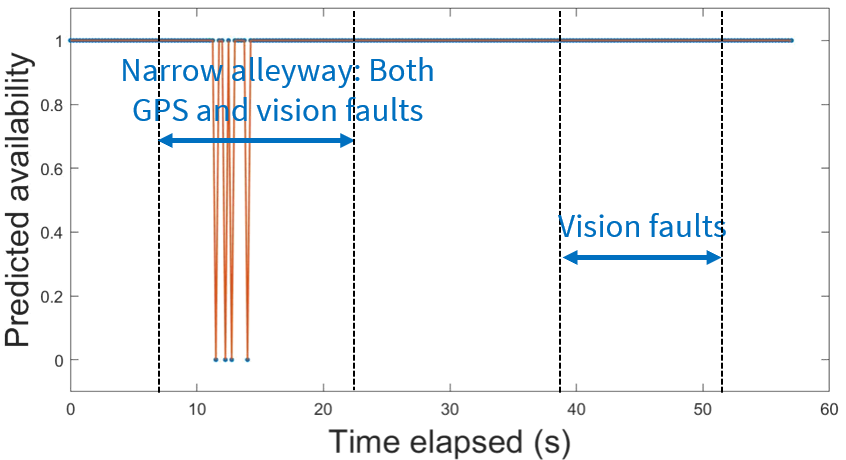}
	\caption{Predicted availability via the proposed ILA demonstrates that the system is available for $98.24\%$ of the entire time. }
	\label{fig_ION:avail}
\end{figure}

In Fig.~\ref{fig_ION:avail}, we demonstrate that the proposed ILA technique provides a robust measure of predicted availability with the system being available for $98.24\%$ of the entire time duration. The predicted availability of $1$ indicates that the worst-case position bounds are less than an $AL=7.5~$m, and $0$ indicates otherwise. In a narrow alleyway, the proposed ILA technique successfully detects degraded localization accuracy, i.e., predicted availability $=0$, due to large faults in both GPS and vision, and is therefore reflective of the measurement quality in urban surroundings. At other time iterations, the proposed ILA technique that predicts the position error bounds performs a robust landmark selection (i.e., among GPS and vision) to ensure compliance with the pre-defined AL.   

\section{CONCLUSIONS} \label{sec_ION:conclusions}
In summary, we developed an Integrity-driven Landmark Attention (ILA) technique for GPS-vision navigation that is inspired by cognitive attention in humans. We perform a two-tiered approach: cost formulation via Stochastic Reachability (SR) and optimization via convex relaxation to select the subset of desired landmark measurements from available, namely GPS satellites and 3D visual features. Given the known measurement error bounds for each landmark in non-faulty conditions, we independently analyze the received measurements to estimate the stochastic reachable set of expected position. By parameterizing the stochastic set via probabilistic zonotope (p-Zonotope), we apply set union property to compute the size of position error bounds as a function of included landmarks. Our ILA technique works with any off-the-shelf GPS-vision estimator and follows a unified approach to account for multiple faults in GPS and vision measurements. 

We validated the proposed ILA technique via an urban dataset with real-world camera images and simulated GPS data. We demonstrated the improved localization accuracy of the proposed ILA technique with maximum position error of only $6.6~$m compared to other techniques for landmark selection that show maximum position errors of $>13~$m, thereby violating the pre-defined alert limit. We also showcase that our ILA technique provides a robust measure of predicted availability with the system being available for $98.24\%$ of the entire duration.

\bibliographystyle{ieeetran}
\bibliography{IEEEabrv,mybiblibraryOne,mybiblibraryTwo}

\end{document}